\documentclass[sigconf]{acmart}

\usepackage{booktabs} 

\setcopyright{rightsretained}

\acmDOI{10.1145/nnnnnnn.nnnnnnn}

\acmISBN{978-x-xxxx-xxxx-x/YY/MM}

\acmConference[GECCO '19]{the Genetic and Evolutionary Computation Conference 2019}{July 13--17, 2019}{Prague, Czech Republic}
\acmYear{2019}
\copyrightyear{2019}

\acmPrice{15.00}

\usepackage{todonotes}

\begin{document}
\title[KL-Div for Game Analysis and PCG]{Tile Pattern KL-Divergence for \\ Analysing and Evolving Game Levels}

\author{Simon M. Lucas}
\affiliation{%
  \institution{Queen Mary University of London, UK}
}
\email{simon.lucas@qmul.ac.uk}

\author{Vanessa Volz}
\affiliation{%
  \institution{Queen Mary University of London, UK}
}
\email{v.volz@qmul.ac.uk}

\renewcommand{\shortauthors}{S. Lucas et al.}

\begin{abstract}

This paper provides a detailed investigation of using the Kullback-Leibler (KL) Divergence as a way to compare and analyse game-levels, and hence to use the measure as the objective function of an evolutionary algorithm to evolve new levels. We describe the benefits of its asymmetry for level analysis and demonstrate how (not surprisingly) the quality of the results depends on the features used. Here we use tile-patterns of various sizes as features.

When using the measure for evolution-based level generation, we demonstrate that the choice of variation operator is critical in order to provide an efficient search process, and introduce a novel convolutional
mutation operator to facilitate this.  We compare the results with alternative generators, including evolving in the latent space of generative adversarial networks, and Wave Function Collapse. The results clearly show the proposed method to provide competitive performance, providing reasonable quality results with very fast training and reasonably fast generation.

\end{abstract}

\copyrightyear{2019} 
\acmYear{2019} 
\setcopyright{acmlicensed}
\acmConference[GECCO '19]{Genetic and Evolutionary Computation Conference}{July 13--17, 2019}{Prague, Czech Republic}
\acmBooktitle{Genetic and Evolutionary Computation Conference (GECCO '19), July 13--17, 2019, Prague, Czech Republic}
\acmPrice{15.00}
\acmDOI{10.1145/3321707.3321781}
\acmISBN{978-1-4503-6111-8/19/07}

%
%

\ccsdesc[500]{Mathematics of computing~Probabilistic representations}
\ccsdesc[500]{Applied computing~Computer games}
\ccsdesc[100]{Theory of computation~Evolutionary algorithms}

\keywords{Kullback-Leibler Divergence, Procedural Content Generation, Latent Vector Evolution, Wave Function Collapse}

\maketitle

\section{Introduction}

Procedural Content Generation (PCG) uses algorithms that to generate content, specifically in the context of games and digital entertainment. The type of content varies widely, including levels, landscapes, narratives and weapons.

A popular subset of PCG techniques is based on searching a space of potential content in order to identify high quality examples. Search-based PCG approaches are especially useful if it is difficult to find an approach that only generates content that satisfies certain constraints, such as e.g. solvability of generated puzzles. In order to apply a search-based approach, three components are needed \cite{searchPCG}:
\begin{itemize}
    \item A searchable space where each point represents a specific instance of generated content;
    \item A search algorithm;
    \item A fitness function to guide the search.
\end{itemize}

However, specifying a suitable fitness function is a difficult problem due to the subjective nature of preferences in games and game design. The functions used in successful PCG approaches are usually very specific to the content they are designed to evaluate (see \cite{Browne-Automatic} for an example).
The ideal measure should correlate with human
preferences in some way, and be fast to compute
in order to provide efficient analysis.  Any such measure also has
the potential to be used as the fitness function for
an evolutionary algorithm and hence be used to generate levels.

A common usecase of search-based PCG is the generation of novel content that is still similar in various aspects to pre-existing, manually designed examples. This usecase is a large part of the motivation behind PCGML approaches (PCG via Machine Learning) \cite{summerville2017procedural}, which revolves around the identification of patterns in data sets. Level generation is a typical example of this usecase. Here, a fitness measures generally express similarity to existing content.

In this paper, we propose to use the Kullback-Leiber Divergence ($D_{KL}$) between generated content and training samples as a measure for similarity in PCGML. The $D_{KL}$ is an asymmetric measure, which enables us to measure and control the degree of novelty in the generated content. We demonstrate how this can be used to gain detailed insights into the behaviour of a given PCG method using Mario level generation as an illustrative example.

After a survey of related work on search-based PCG in section \ref{sec:pcg}, we formally introduce the $D_{KL}$ measure and its adaptation in the context of Mario levels in section \ref{sec:kldiv}. We show that the $D_{KL}$ correlates 
at least in some way with human intuition of level similarity in section \ref{sec:kldiv:eval}, thus validating the measure for the purposes of this paper. Following that, in section \ref{sec:kldiv:gen} we present an efficient level generation method called ETPKLDiv (Evolution using Tile-Pattern KLDiv) based around $D_{KL}$, which successfully creates Mario levels in the style of existing ones. We compare the levels with others generated from Wave Function Collapse (WFC, see section \ref{sec:rw:wfc}) and Latent Vector Evolution (MarioGAN, see section \ref{sec:rw:mariogan}), two state-of-the-art PCG methods. The evaluation of the results can be found in section \ref{sec:gen:results}. Finally, we summarise our results and give an outlook on future work in section \ref{sec:final}.

We find that ETPKLDiv offers competitive performance in terms of speed of training, speed of generation and generalisation from small samples.  Solution quality is harder to assess but the method is able to evolve reasonable looking levels that often satisfy the objective function better snippets of the training level.



\section{Search-Based Procedural Content Generation}
\label{sec:pcg}

Although there has been significant work on many aspects of
PCG, the viral success of Wave Function Collapse\footnote{\url{https://github.com/mxgmn/WaveFunctionCollapse}} \cite{karth2017wavefunction} 
demonstrates the appetite for PCG which can rapidly
generate interesting content from small amounts of training data.

When dealing with search-based PCG there are several
(not mutually exclusive) approaches that can be applied
to evaluate candidate solutions as identified in various taxonomies (e.g. \cite{pcgbook2}):

\begin{itemize}
    \item Use some hand-designed criteria. These tend to be specific to a game or game genre;
        \item Use agent-based play testing. This approach has become more attractive with the development of general game AI based on statistical forward planning algorithms such as Monte Carlo Tree Search \cite{browne2012survey} and Rolling Horizon Evolution \cite{perez2013rolling};
        \item Use game-level samples (also referred to as 
Procedural Content Generation via Machine Learning (PCGML) \cite{summerville2017procedural}.
\end{itemize}

In this paper, we are specifically targeting PCGML.

Several methods have been proposed in order to evaluate content generators (see overview in \cite{pcgbook12}). They mostly revolve around the characterisation of generated content according to features such as novelty and difficulty. Some of these methods are more explorative (e.g. \emph{expressive range}), where others are mainly focused on optimisation, as in this paper.

There is a wide range of quality indicators, but for the PCGML usecase targeted in this paper, we specify the following objectives:


\begin{itemize}
    \item The quality of the generated content.  Some aspects of quality can be quantified while others are inherently subjective;
    \item The amount of sample data needed for adequate performance;
    \item The flexibility of the generator: e.g.\ can it directly produce levels of any size? Can it produce consistent results for different types of content?
    \item The time taken to train the system;
    \item The time taken to generate each new level once trained.
\end{itemize}


For the purpose of demonstration in this paper, we use the generation of Mario levels as an illustrative example. We introduce this application in section \ref{sec:problem}, along with corresponding related work. Afterwards, we present two state-of-the-art PCG approaches in more detail, as their results are used as a comparison for the algorithm presented in this paper. We describe Wave Function Collapse in section \ref{sec:rw:wfc} and MarioGAN in section \ref{sec:rw:mariogan}.

\subsection{PCGML Mario Level Generation}
\label{sec:problem}
Super Mario Bros. is a typical platformer game, where the main character, Mario, needs to traverse multiple different levels and defeat enemies to win the game. Mario levels are usually indicated to belong to one of five types (overworld, underground, athletic, castle, underwater)\footnote{\url{https://www.mariowiki.com/Super_Mario_Bros.\#List_of_levels}}. The type determines the aesthetic style of the level, but also affects the challenges that are posed. For example, athletic levels usually contain platforms that are spaced far apart, as well as moving platforms. They thus require precise and timed jumps to traverse.

\begin{figure*}
    \centering
    \includegraphics[width=\linewidth]{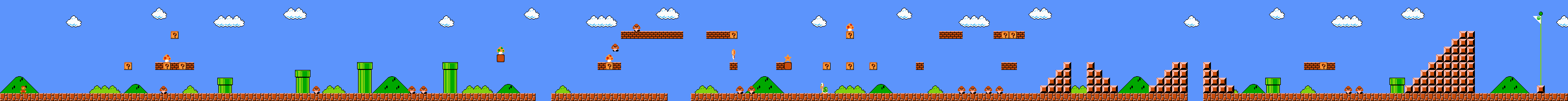}
    \caption{\label{fig:1-1}Super Mario Bros. Level 1-1}
\end{figure*}

In Super Mario Bros., the levels are also grouped into different \emph{worlds}. Levels are named according to which world they belong to (1-8) and based on their order within the world (1-4). For example, the first level in the first world is called 1-1 (depicted in figure \ref{fig:1-1}).
As can be seen in the image, Mario levels consist of several different tiles that are arranged in a 2D grid to form structures such as platforms, stairs and pipes. The full set of tiles contained in Level 1-1 is listed in table \ref{tab:tiles}.

\begin{table}
\centering
\caption{\label{tab:tiles}Tile types in Mario levels. \normalfont 
The symbol characters from the VGLC encoding, and the
numeric identities that are mapped to the corresponding tile types as implemented in the Mario AI framework to produce the visualization shown. Table reproduced from \cite{volz:gecco2018}}
\begin{tabular}{cccc}
\hline
Tile type & Symbol & Identity & Visualization\\
\hline 
Solid/Ground & X & 0 & \includegraphics[scale=0.5]{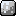}\\
Breakable & S & 1 & \includegraphics[scale=0.5]{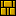}\\
Empty (passable) & - & 2 & \\
Full question block & ? & 3 & \includegraphics[scale=0.5]{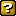}\\
Empty question block & Q & 4 & \includegraphics[scale=0.5]{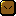}\\
Enemy & E & 5 & \includegraphics[scale=0.5]{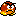}\\
Top-left pipe & < & 6 & \includegraphics[scale=0.5]{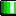}\\
Top-right pipe & > & 7 & \includegraphics[scale=0.5]{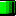}\\
Left pipe & [ & 8 & \includegraphics[scale=0.5]{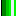}\\
Right pipe & ] & 9 & \includegraphics[scale=0.5]{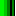}\\
Coin & o & 10 & \includegraphics[scale=0.5]{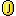}\\
\hline
\end{tabular}
\end{table}

The table also contains a list of symbols that are used to represent the respective tile type in the Video Game Level Corpus (VGLC) \cite{Summerville:pcg2016-VGLC}. The VGLC contains a collection of levels from tile-based games in such an encoding. The Super Mario Bros. levels from the VGLC serve as the sample set for the illustrative usecase in this paper.

Specifically, level generators have to produce fixed-size matrices using the encoding from VGLC as described in table \ref{tab:tiles}. In order to keep consistent with existing work on PCGML Mario Level generation, only level 1-1 is used as a training set for the generators.  Since PCGML aims to learn from the training data, the generated
levels should be similar to the original. Just how similar they should be is a moot point, since the aim is to generate content which is in some way novel while looking and playing like the training sample. Here, we use KL-Divergence to measure similarity, as described in detail in section \ref{sec:kldiv}. We validate this decision in section \ref{sec:kldiv:eval}.

\subsection{Wave Function Collapse}
\label{sec:rw:wfc}

Wave Function Collapse is a PCG approach that has gained recent prominence, especially from its application in popular games such as Bad North\footnote{https://www.badnorth.com/} (Raw Fury, 2018). The approach relies on fitting patterns together such that a set of constraints are fulfilled \cite{karth2017wavefunction}. These constraints are in the format of which patterns fit next to each other given an offset. Constraints are either handcrafted, or learned from a set of training samples.

After identification of patterns and constraints (1), all valid patterns that are still eligible for placement can thus be identified for each open spot (2). Next, the open spot with the minimal entropy of eligible patterns is identified (3) and filled using a pattern randomly selected from the set of available patterns for this particular spot (4). The random selection is often biased by the frequency of the patterns in the training sample. Steps 2-4 are repeated until either all spots are filled, or until the constraints contradict.

The description of the algorithm above as well as its implementation for this paper are based on \cite{karth2017wavefunction}.

\subsection{MarioGAN}
\label{sec:rw:mariogan}
Another recently proposed PCG approach applies the concept of Generative Adversarial Networks (GANs) to creating content for games. The approach further employs a latent vector search and is thus called Exploratory Latent Search GAN (ELSGAN) within the context of this paper. It was applied to Mario levels in \cite{volz:gecco2018}.

The generator in ELSGAN is a multi-layer neural network that, based on an input vector $[-1,1]^{32} \in \mathbb{R}$, outputs several real-valued vectors that can be translated into Mario levels using a one-hot encoding. The generator is trained using an adversarial neural network, often dubbed the discriminator \cite{arjovsky2017wasserstein}. During training, the discriminator is tasked with identifying whether levels presented to it are created by the generator, or are part of the set of original training samples. The discriminator then receives feedback based on the error rate in this classification task. In contrast, the generator is rewarded if it is able to "fool" the generator. Ideally, after many iterations of alternating training of generator and discriminator, the system converges at a point where the generator produces outputs that are indistinguishable from the training samples.

\section{KL-Divergence as fitness measure}
\label{sec:kldiv}

The Kullback-Leibler Divergence $D_{KL}$, also called relative entropy, is the expectation
of the logarithmic differences between two probability distributions $P$ and $Q$ where the expectation (weighted sum)
is calculated using the probability distribution $P$.\footnote{The $D_{KL}$ is only properly defined if $\forall x$, $Q(x)=0~\implies~P(x)=0$.}
\begin{equation}
    D_{KL}(P||Q) = \sum_{x \in X} P(x) \mbox{log} \left ( \frac{P(x)}{Q(x)} \right )
    \label{eq:KLDiv}
\end{equation}

Compared to \emph{ad hoc} approaches to comparing levels such as taking the absolute difference between feature histograms, the $D_{KL}$ has two practical advantages: its asymmetry can be used to good effect, and the probabilistic nature means proper weight is given to differences in feature counts.  For example, the difference between $0$ and $1$ occurrences is weighted much higher than the difference between $10$ and $11$ occurrences.

\begin{figure}[hbtp]
\centering
\includegraphics[width=0.85\linewidth]{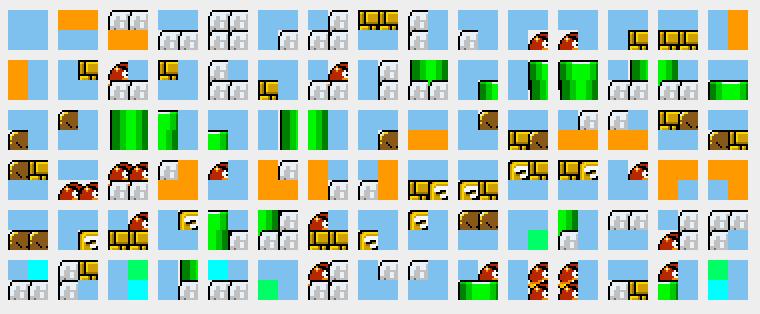}

    \caption{\label{fig:2x2-tiles}
    The 90 distinct $2x2$ tile patterns in the training level,
    shown from most to least frequent.  Top left pattern (plain blue sky) occurs
    $2,100$ times.
    }

\end{figure}

\begin{figure}[hbtp]
\centering
\includegraphics[width=0.85\linewidth]{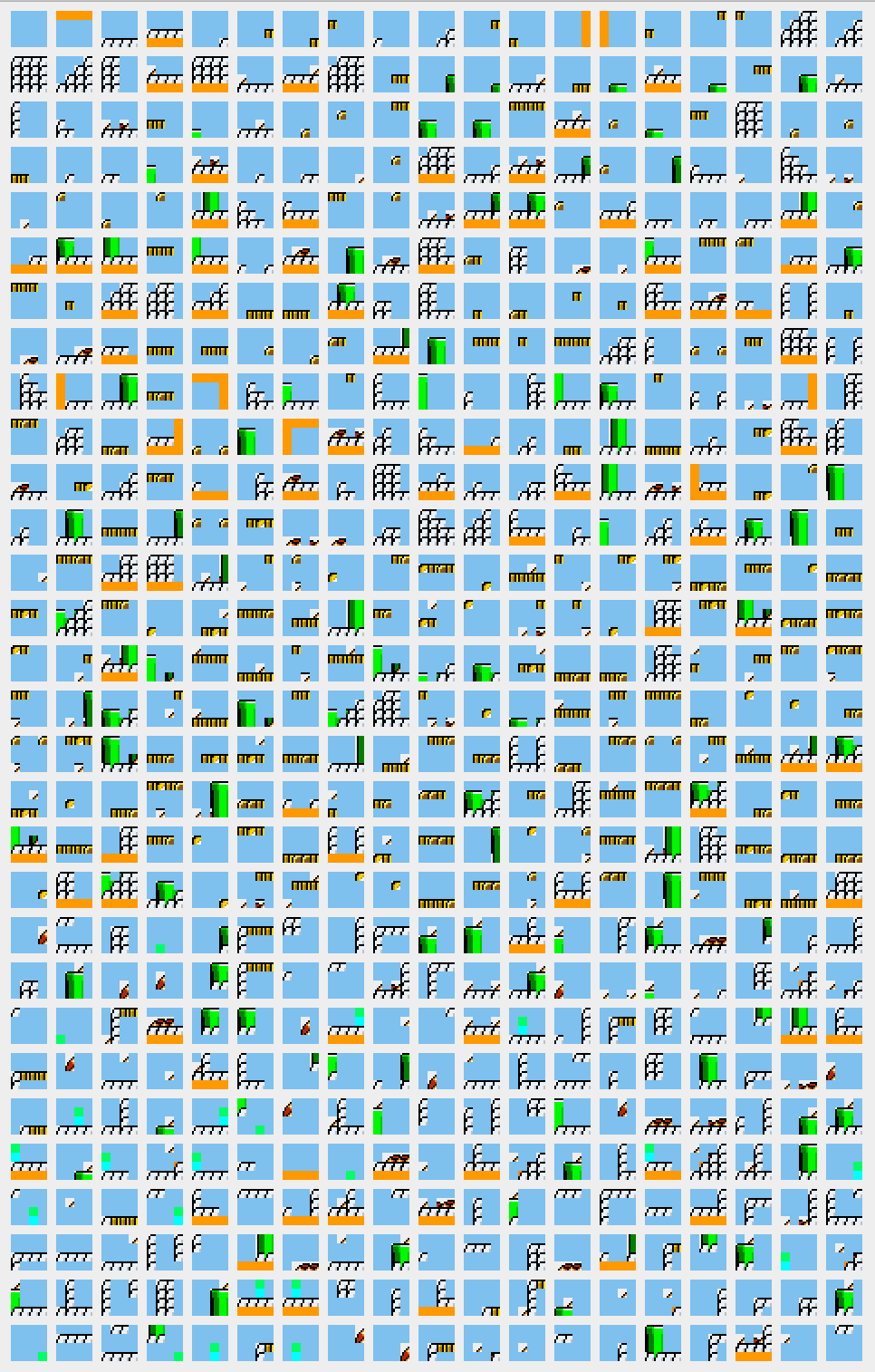}

    \caption{\label{fig:4x4-tiles}
    The 570 distinct $4x4$ tile patterns in the training level, shown 
    from most to least frequent.  Top left pattern occurs 1,349 times
    whereas the least frequent 50\% of the patterns occur only once
    or twice.
    }

\end{figure}

In this paper, we define $D_{KL}$ for Mario levels (see section \ref{sec:problem}). In order to apply the $D_{KL}$, we transform levels into probability distributions over tile pattern occurrences. Given a rectangular level of size $(T_W \times T_H)$ and a rectangular filter window of size $(F_W \times F_H)$, the total number of tile patterns $N$ is given by:
\begin{equation}
    N = (1 + T_W - F_W)(1 + T_H - F_H)
\end{equation}
Thus, $X$ in equation \ref{eq:KLDiv} is the set of all tile patterns observed in either the set of training samples, or in the generated levels. For each experiment, we take the set of tile patterns that occur by sliding (convolving) a fixed-size window over a level, where a level is defined as a 2-d array 
of tiles (see table \ref{tab:tiles}). 
The training sample shown in figure~\ref{fig:1-1}
has $90$ distinct $2\times2$ patterns and $570$ 
distinct $4x4$ patterns, shown in Figures
~\ref{fig:2x2-tiles} and~\ref{fig:4x4-tiles},
respectively.

We estimate the probability of a tile pattern using a frequentist approach i.e. we count the number of times pattern $x$ occurs in a sample and divide by the total number of tile occurrences (i.e. by $|X|$). When calculating $D_{KL}$ we ignore any patterns that occur in $Q$ but not in $P$, because if $P(x)=0$, the corresponding summand also equals $0$. However, we add a small constant $\epsilon$ to each probability estimate for $Q(x)$ to avoid divide by zero errors. This also accounts for the fact that, just because a pattern was never observed, does not mean that its probability is truly zero - it just means it has not been observed yet (similar methods are used when building statistical models of natural language).  

Let $X_P$ be the set of patterns observed in probability distribution $P$, and $P'(x)$ and $Q'(x)$ be the epsilon-corrected probability estimates. Further, let $C(x)$ be the number of occurrences of $x$ in a sample $X$ and $C=\sum_{x \in X_P} C(x)$ be the sum of $C(x)$ over all $x \in X$. We then we compute $P'(x)$ as follows:
\begin{equation}
P'(x) = \frac{C(x) + \epsilon}{(C+\epsilon)(1+\epsilon)}
\label{eq:SafeP}
\end{equation}

Based on this, our safe approximation of the $D_KL$ is:
\begin{equation}
    D_{KL}(P||Q) = \sum_{x \in X_P} P'(x) \mbox{log} \left (\frac{P'(x)}{Q'(x)} \right )
    \label{eq:KLDivSafe}
\end{equation}


$D_{KL}(P||Q)$ is always greater than or equal to zero,
and only zero if $P$ is identical to $Q$.
Since $D_{KL}$ is an asymmetric measure, we have 
to decide which way around to apply it, or whether to
apply it symmetrically by adding the two results together.
Here we choose a weighted approach that can vary smoothly
between the two extremes. We transform the resulting function into a maximisation problem by negating the $D_{KL}$.
Hence we define the fitness
of a generated level $Q$ with respect to a set of sample
levels $P$ as:

\begin{equation}
    F(P,Q) = - \left( w \cdot D_{KL}(P||Q) + (1-w) \cdot D_{KL}(Q||P) \right)
    \label{eq:Fitness}
\end{equation}

\subsection{Parameters}
\label{sec:kldiv:param}
The behaviour of $F(P,Q)$  is characterised by four parameters:

\begin{itemize}
\item $\epsilon$ is the back-off estimation constant (see equation \ref{eq:SafeP}).
\item Dimensions of filter: width $F_w$ and height $F_h$ in tiles.  
\item The weight $w$ used to balance between the two asymmetric $D_{KL}$ terms (see equation \ref{eq:Fitness}).
\end{itemize}

As we shall see in section~\ref{sec:asymmetry}, $w$ can be used to good effect to
control the degree of novelty in the generated levels.
Setting $w=0$ aims for no novelty in the tile patterns i.e. all tile patterns in the generated level should have occurred at least once in
the training sample.  Note that the arrangement of tile-patterns is still likely to be novel.
Satisfying this constraint may be at the expense of 
missing out on many patterns that did occur.  
On the other hand, setting $w=1$
favours trying to include all patterns observed in
the training sample in the generated level: this may be impossible if the
dimensions of the generated level are smaller than the training sample,
which is commonly the case.

\subsection{Evaluation and Validation}
\label{sec:kldiv:eval}

The Kullback-Leibler Divergence (denoted as $D_{KL}$ here) has been used successfully for Mario level generation in the context of Mario\-GAN \cite{volz:gecco2018}. This might not be obvious immediately, but it was shown in \cite{goodfellow2014generative} that the training process for generative adversarial networks essentially minimises the $D_{KL}$ between the distributions of the generator output and the original training samples. Hence, both the method used in this paper and the Mario level generation from GANs in \cite{volz:gecco2018} use the concept of KL-Divergence, but apply it in different ways. Here, we apply it to tile-pattern distributions, whereas GANs are computing it over individual tile-positions, but from a latent encoding that forces the tiles into structured patterns.

Below, we validate its usage as a fitness measure for generated levels. The main question is whether $D_{KL}$ is able to correctly identify patterns in 2D tile-based levels. To this end, we conduct the following analysis using hierarchical clustering with average linkage. We test whether the $D_{KL}$ is able to correctly identify the similarities between levels from one type and the differences to others.

\begin{figure}
    \centering
    \includegraphics[width=0.9\linewidth]{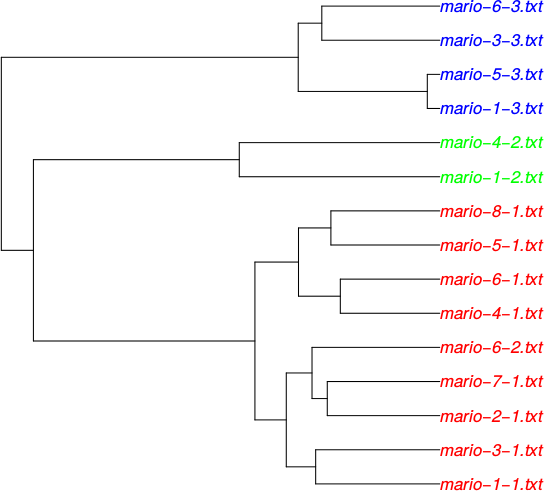}
    \caption{\label{fig:cluster}Dendrogram for clusters based on $D_{KL}$  with filter $F_w \times F_h = 4\times4$ and weight $w=0.5$. Colours denote level type (overworld in red, underground in green, athletic in blue).}
\end{figure}

We compute a similarity matrix using $D_{KL}$ for all Super Mario Bros. levels contained in the VGLC \cite{Summerville:pcg2016-VGLC}. This set of levels only contains overworld, underground and athletic levels (see section \ref{sec:problem} for details). Separate experiments were conducted for $D_{KL}$ measures based on various small filter sizes $F_w$ and $F_h$ between $1$ and $5$, and for weights $w \in \{0,0.5,1\}$. In all our experiments, three distinct clusters were detected, corresponding to the 3 different types of levels. A dendrogram of these results is shown in figure \ref{fig:cluster} for filter size $F_w \times F_h = 4\times4$. We have thus demonstrated that $D_{KL}$ is useful to identify similarities and dissimilarities in Mario levels.

Besides this rudimentary validation of the expressiveness of the measure, a  benefit of the tile-pattern $D_{KL}$ measure is its potential for further analysis of the generator and the generated content. For example, based on the contribution of singular patterns to the $D_{KL}$, we are able to identify anomalies. Furthermore, the $D_{KL}$ is able to compress information from tile-pattern occurrence histograms into a single meaningful number. By changing the filter sizes, the measure is also able to express similarity on different levels of granularities.


\section{KL-Divergence Generator}
\label{sec:kldiv:gen}

Using $D_{KL}$ as the basis of a procedural level generator is conceptually 
simple but achieving good results requires some careful design. For the experiments in this paper, we used a Random Mutation Hill Climber (also known as a (1+1) EA). This simple algorithm has shown competitive performance across a range of problems
such as evolving finite automata \cite{LucasDFA}.

However, the tile-pattern  $D_{KL}$ fitness function induces a rugged fitness landscape 
when using a standard mutation operator because many parts of a solution have to be changed in a coordinated way in order to improve the solution. This problem is exacerbated for larger window sizes. In the next subsections we will first
show examples evolved levels to illustrate the problem and then
introduce a convolutional mutation operator that enables the $(1+1)$ EA
to achieve acceptable performance.  We judge it as acceptable in
two ways: (1) the fitness values are often better than the same size snippets 
from the sample level, and (2) in our opinion they look like reasonable Mario levels.

\subsection{Standard Mutation Operator}

\begin{figure}[hbtp]
\centering
\includegraphics[width=0.8\linewidth]{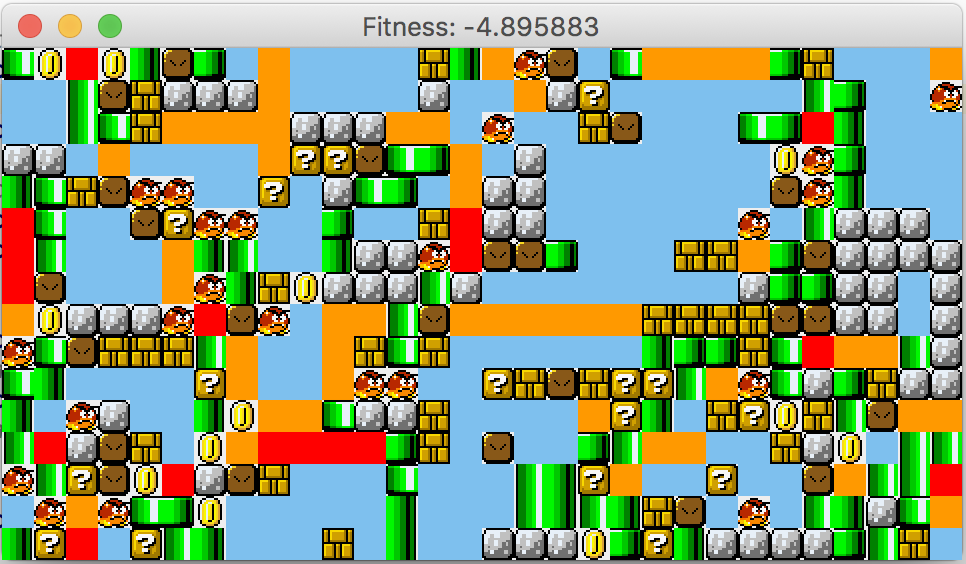}
\caption{\label{fig:Evolved-2x2-Regular}
Sample level generated after evolving for $10,000$ fitness evaluations using a $2\times2$ filter
and a standard mutation operator.
}
\end{figure}

\begin{figure}[hbtp]
\centering
\includegraphics[width=0.8\linewidth]{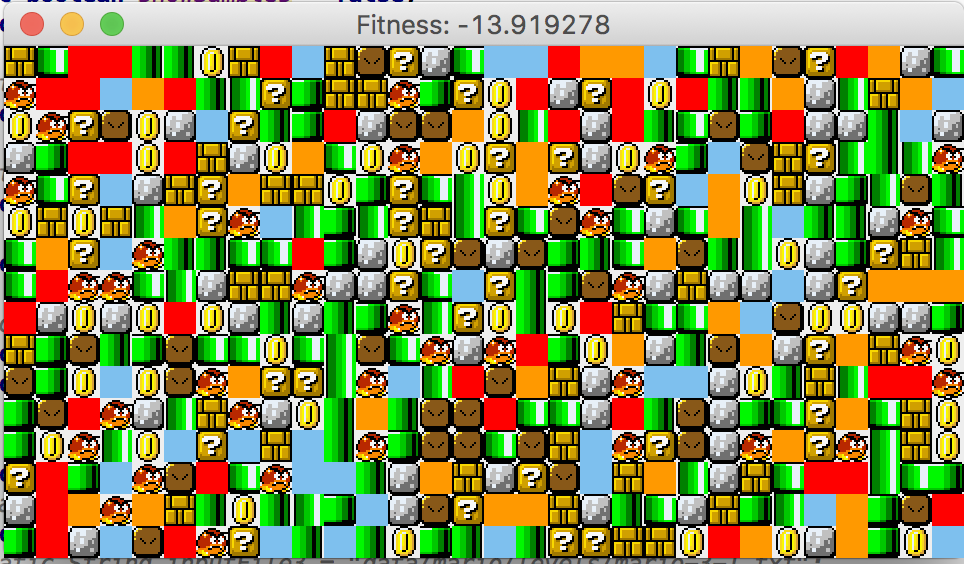}
\caption{\label{fig:Evolved-4x4-Regular}
Sample level generated after evolving for $10,000$ fitness evaluations using a $4 \times 4$ filter and a standard mutation operator.  The search gets nowhere and this solution
has similar fitness to a randomly generated level.
}
\end{figure}

Figure~\ref{fig:Evolved-2x2-Regular} shows a level evolved using a standard ``flip'' mutation operator.  This operator works by scanning each tile in the level, changing it to a randomly selected tile with a given (typically small) probability. We set this to on average flip 3 tiles per application of the operator.  A typical result of evolving using a $2x2$ filter window for $100,000$ fitness evaluations is shown in Figure~\ref{fig:Evolved-2x2-Regular}. Clearly, progress has been made compared to a uniform random level, but there are still many anomalies compared to the sample level.  The regular mutation operator fails badly with larger windows such as $3 \times 3$ or bigger. Figure~\ref{fig:Evolved-4x4-Regular} shows an example using a $4 \times 4$ filter which fails to improve over a random point in the search space after 100,000 iterations (the figure is therefore an example of a random level).

\subsection{Convolutional Mutation Operator}

The problem with the standard mutation operator 
is that it is unlikely that any number of
randomly flipped tiles will lead to an improvement
in fitness.  For example, improving the fitness will often involve modifying a sub-rectangle of the level so that non-matching windows will be transformed into ones that do match at least one pattern observed in the training sample.  Typically, several tiles may need to be flipped in order to make a single additional match.  However, many changes may also destroy existing matches with overlapping windows which further compounds the problem.  To counter this, we designed a novel mutation operator which is both simple and effective.  The operator samples a filter-size rectangle from the training set and copies it to a random location in the generated level.  This now has a much higher chance of leading to an improvement in fitness.  

Indeed, experiments show that using this mutation operator we can evolve fitter levels of a particular size (e.g. of width 30 tiles) than are present in the training data (as measured by fitness function $F$). This is because the training tile distribution is drawn from a single wide level that progresses through various phases, each of which tends to favour or exclude particular pattern types. The optimiser then tries to squeeze them all into the available space. If desired, similar levels of tile densities can be achieved by simply increasing the width of levels to be generated, though larger levels lead to slower convergence.

\begin{figure}[hbtp]
\centering
\includegraphics[width=0.8\linewidth]{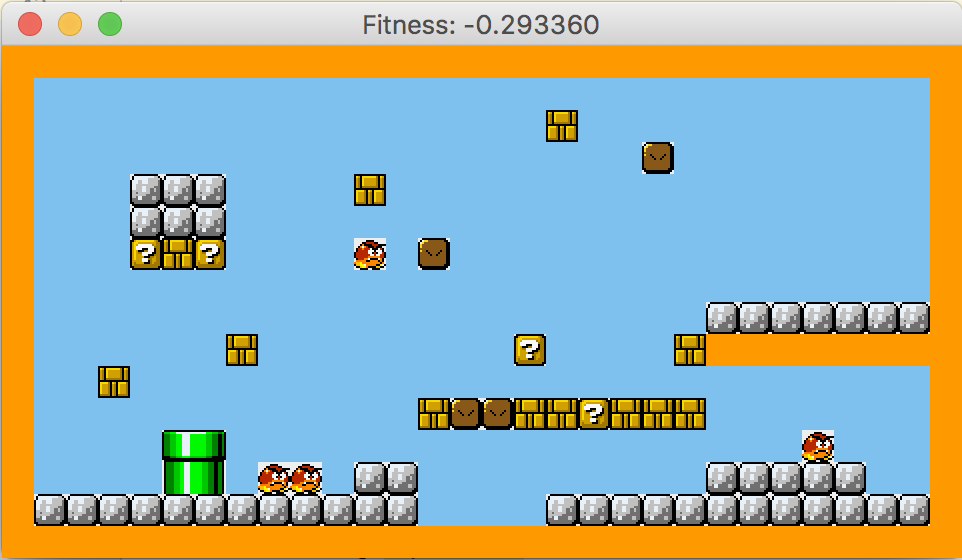}
\caption{\label{fig:Evolved-2x2-Conv}
Sample level generated after evolving for $10,000$ fitness evaluations using a $2\times2$ filter
using a convolutional mutation operator.
}
\end{figure}

\begin{figure}[hbtp]
\centering
\includegraphics[width=0.8\linewidth]{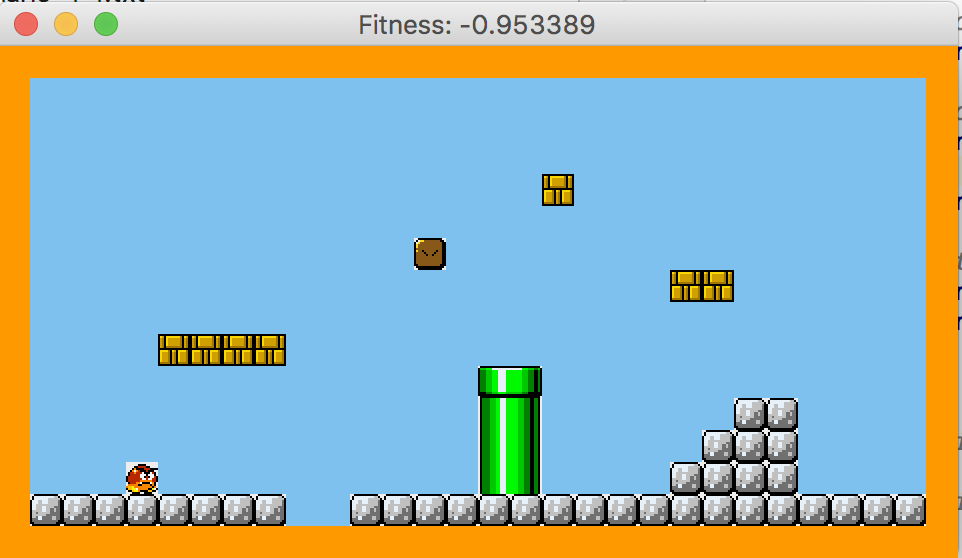}
\caption{\label{fig:Evolved-4x4-Conv}
Sample level generated after evolving for $10,000$ iterations using a $4\times4$ filter
using a convolutional mutation operator.  Note how this is more similar to the training
sample now, and yet still has some novelty.
}
\end{figure}

Figures~\ref{fig:Evolved-2x2-Conv} and \ref{fig:Evolved-4x4-Conv}
show levels evolved using $2 \times 2$ and $4 \times 4$
convolutional mutation operators.  The $2 \times 2$ example
almost looks like a reasonable level but has some anomalies,
whereas the $4 \times 4$ example is a fine looking level.

The evolutionary generator takes just under 1 second for $10,000$ fitness
evaluations using a $2 \times 2$ filter and just under 2 seconds when
using a $4 \times 4$ filter.

\begin{figure}[hbtp]
\centering
\includegraphics[width=0.95\linewidth]{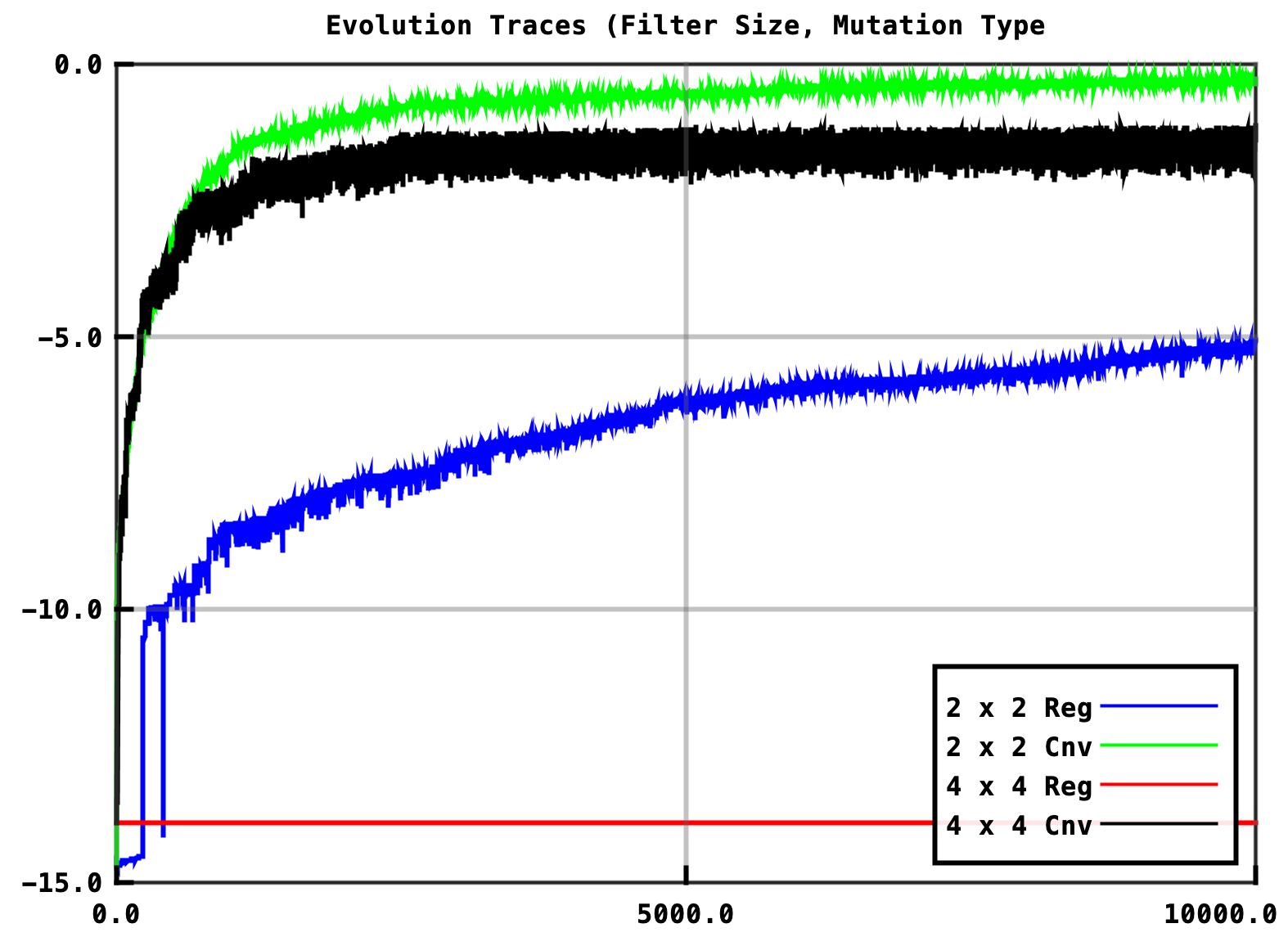}
\caption{\label{fig:EvoTraces}
Evolutionary traces (plots of the fitness for each sampled candidate for evolutionary runs of 10,000 evaluations) for four combinations of filter size and 
mutation operator.  Note that using a $4 \times 4$ filter
with a regular mutation operator makes no progress at all.
}
\end{figure}

Figure~\ref{fig:EvoTraces} shows typical evolutionary
traces using different filter size and mutation 
operator combinations, each time running the $(1+1)$ EA for
$10,000$ fitness evaluations.  The graph shows the fitness
$F$ of each sample for each combination.  Larger filter sizes
lead to more constraints due to the greater overlap of each filter 
position.  Evolving with the smaller filter size therefore
leads to more rapid improvements and better final fitness,
at the expense of producing less satisfactory levels with
more anomalies.

\subsection{Generation  from Small Samples}

\begin{figure}[hbtp]
\centering
\framebox{\includegraphics[width=0.18\linewidth,trim=4 4 4 4]{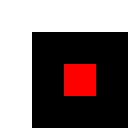}}
\caption{\label{fig:TinySample}
Small $4 \times 4$ image patch.
}
\end{figure}

\begin{figure}[hbtp]
\centering
\includegraphics[width=0.8\linewidth]{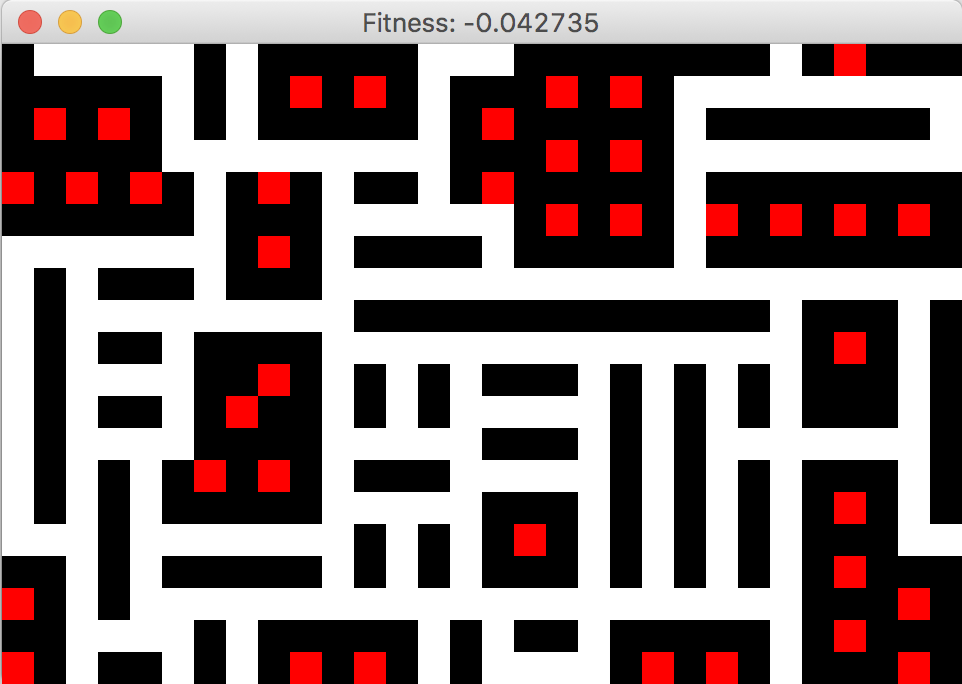}
\caption{\label{fig:EvoFromTiny}
Sample tile image generated after evolving for $10,000$ iterations using a $2\times2$ filter
from the $4 \times 4$ image patch in figure \ref{fig:TinySample}
using a convolutional mutation operator.  Unlike the WFC
approach, this contains tile patterns that do not occur in the sample.
}
\end{figure}

One of the challenges mentioned by the PCGML survey \cite{summerville2017procedural} was
generating content from small samples.  One of the appeals of WFC is its ability
to do this.  To this end we also tested our evolutionary method on one
of the tiny samples in the WFC repository, where a single $4 \times 4$ tile pattern (shown in figure~\ref{fig:TinySample})
is used to generate interesting levels of arbitrary size.  Here we used
a $2 \times 2$ filter, and $10,000$ fitness evaluations for the evolution.  Figure~\ref{fig:EvoFromTiny} shows
a sample tile image generated from this tiny pattern
demonstrating that as with WFC, interesting images can 
be built from tiny samples.  Unlike WFC, our
method may also introduce novel sub-patterns, whether desired
or not.

\subsection{Exploring KL-Div Asymmetry}

\label{sec:asymmetry}

\begin{figure}[hbtp]
\centering
\includegraphics[width=0.9\linewidth]{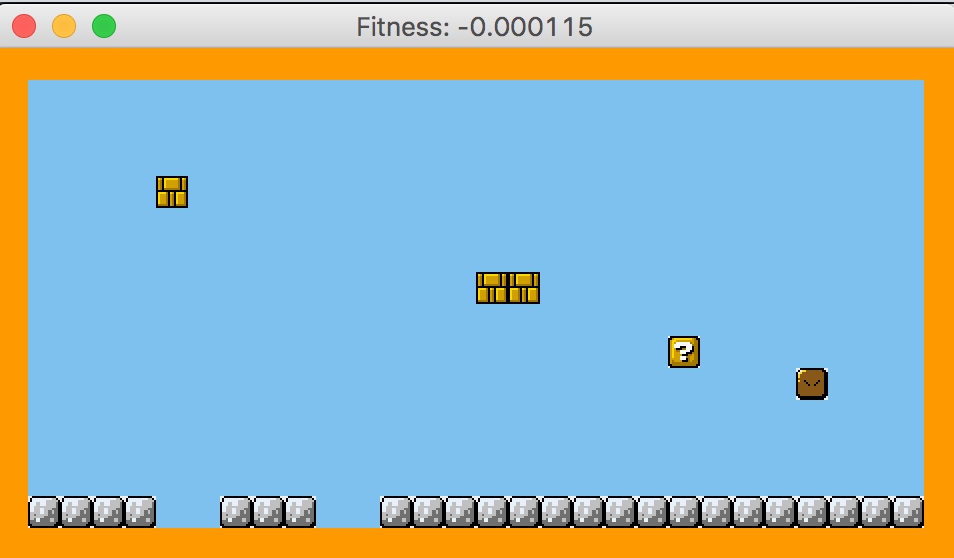}
\caption{\label{fig:PunishNovelty}
Level generated using asymmetric KL-Div ($w=0.0$) where novel patterns (those not occurring in training sample are
penalised, but there is no direct penalty for not using sample patterns.
}
\end{figure}

Here we explore the asymmetric nature of $D_{KL}$ .  If we measure 
the divergence of the generated level from the training sample $(w=0.0)$
we get very different results compared to making the opposite
calculation.  Figure~\ref{fig:PunishNovelty} shows the effect 
of evolving a level using a $4x4$ filter (our default filter size)
using fitness measure $F$ with $w=0.0$.  
Note how this generates boring levels.
Every tile pattern seen in the level has occurred during training,
but many interesting ones have been omitted because there was insufficient
incentive to include them.  Hence this leads to a ``lazy'' effort of placing lots
of easily placed patterns, leading to excessive sky when generating Mario levels.

\begin{figure}[hbtp]
\centering
\includegraphics[width=0.9\linewidth]{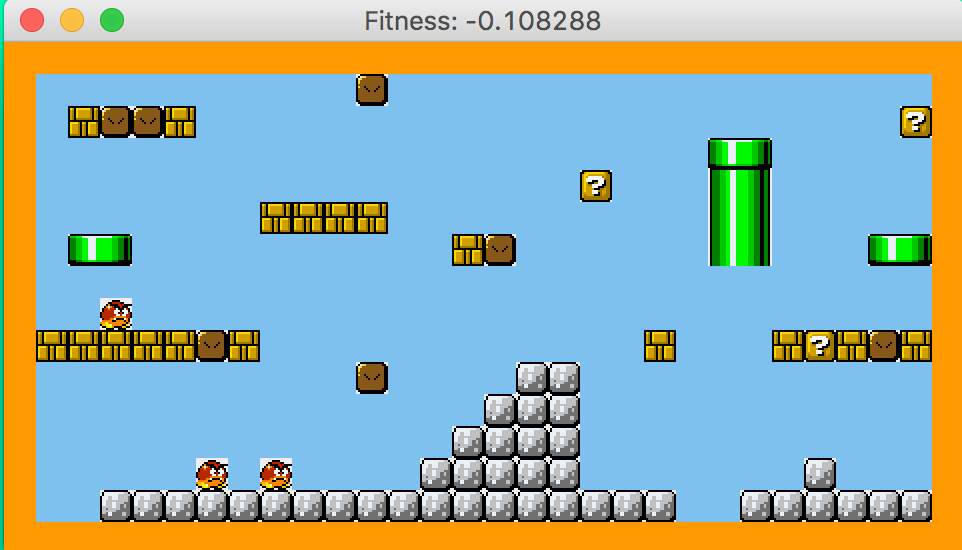}
\caption{\label{fig:PunishExclusion}
Level generated using asymmetric KL-Div ($w=1.0$) where failure to use sample patterns
is penalised, but any novel (unseen) patterns are not penalised.
}
\end{figure}

Observe the contrast in Figure~\ref{fig:PunishExclusion} that
used $(w=1.0)$.  Here we see that many interesting tile
patterns have been included from the training sample, but put together
in ways that also lead to many patterns which never occurred in the 
training sample.

Our default is to use the symmetric KL-Div $(w=0.5)$, which evenly weights the
the two asymmetric measures.  Note that any weighted sum could
be used to favour inclusion of sample patterns or exclusion of 
novel patterns.  We have not tuned this thoroughly and used $(w=0.5)$ for most of
the generated levels in the paper, though values of $0.6$ or $0.7$ may produce even more
interesting looking levels (in our opinion) at the expense of the occasional anomaly.

\subsection{Evaluation}
\label{sec:gen:results}





Table~\ref{tab:comparison} provides a summary of how
the proposed method compares with other recent sample-based methods.  As mentioned above, we include evaluations of Wave Function Collapse (WFC), Evolution in the Latent Space of Generative Adversarial Networks (ELSGAN)\footnote{\url{https://github.com/schrum2/GameGAN}} \cite{volz:gecco2018} and
our proposed method: Evolution with Tile Pattern KL-Divergence (ETPKLDiv). We also include a GAN model without latent vector evolution (GAN). These are not meant to be an exhaustive list of general sample-based 2D generators, but both WFC and ELSGAN have gained recent prominence.
All these methods potentially generalise to any 2D tile-based 
level (hence one-dimensional Markov chain methods are not included).

\begin{table}
\caption{A qualitative rating of the training and generation speed of each method described in this paper.  For MarioGAN we decompose the rating in to the GAN part and the GAN combined with evolution (ELSGAN). \emph{Tiny} refers to whether the method is able to generate whole levels from a single small patch (e.g. see figures \ref{fig:TinySample} and \ref{fig:EvoFromTiny} ).
\label{tab:comparison} }
\begin{tabular}{c|c|c|c}
Method          & Training & Generation & Tiny \\ \hline
ETPKLDiv     & Fast          &  Fast to Medium, Never Fails & Yes  \\
WFC           &   Fast  &  Fast to Slow, May Fail & Yes \\  
GAN        &  Slow   &  Always Fast, Never Fails & No \\
ELSGAN        &  Slow   &  Slow, Never Fails & No \\
\end{tabular}
\end{table}


In table~\ref{tab:comparison} by fast we mean sub-second time, medium is
the order of a few seconds, slow is of the order of minutes, hours or more.
WFC has highly variable generation time compared to the other
methods, and is the only one which can fail to produce a level.
The other methods may produce poor levels (as in ones which 
differ too much from the training sample) but always produce something. Furthermore, the use of tile-pattern features together with $D_{KL}$
leads to flexible analysis and generation, 
since the sliding window means the levels we generate
can be of a different size compared to the training sample(s). This is in contrast
to the MarioGAN method where once trained, the GAN outputs levels of a fixed size.


As we have established in section \ref{sec:kldiv:eval}, $D_{KL}$ is a form of valid evaluation to assess the similarity of generated levels to the original. For each of the approaches in table \ref{tab:comparison}, we thus train a generator from Mario level 1-1 (see figure \ref{fig:1-1}) and generate $100$ levels each. For the approaches using tile patterns (WFC and ETPKLDiv), we test several versions with filter sizes ($2\times2$, $3\times3$ and $4\times4$). For the resulting levels, we then compute the $D_{KL}$ values.

In order to test the effects of the parameters discussed in section \ref{sec:kldiv:param}, we vary the filter sizes and weights. We test all combinations of $2\times2$, $3\times3$ and $4\times4$ with weights in $0, 0.5$ and $1$. These weights represent the symmetric ($0.5$) and extreme cases of asymmetry (see section \ref{sec:asymmetry}).

\begin{figure}
    \centering
    \includegraphics[width=0.4\linewidth]{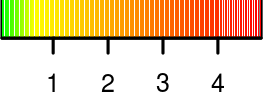}\\
    \includegraphics[width=0.9\linewidth]{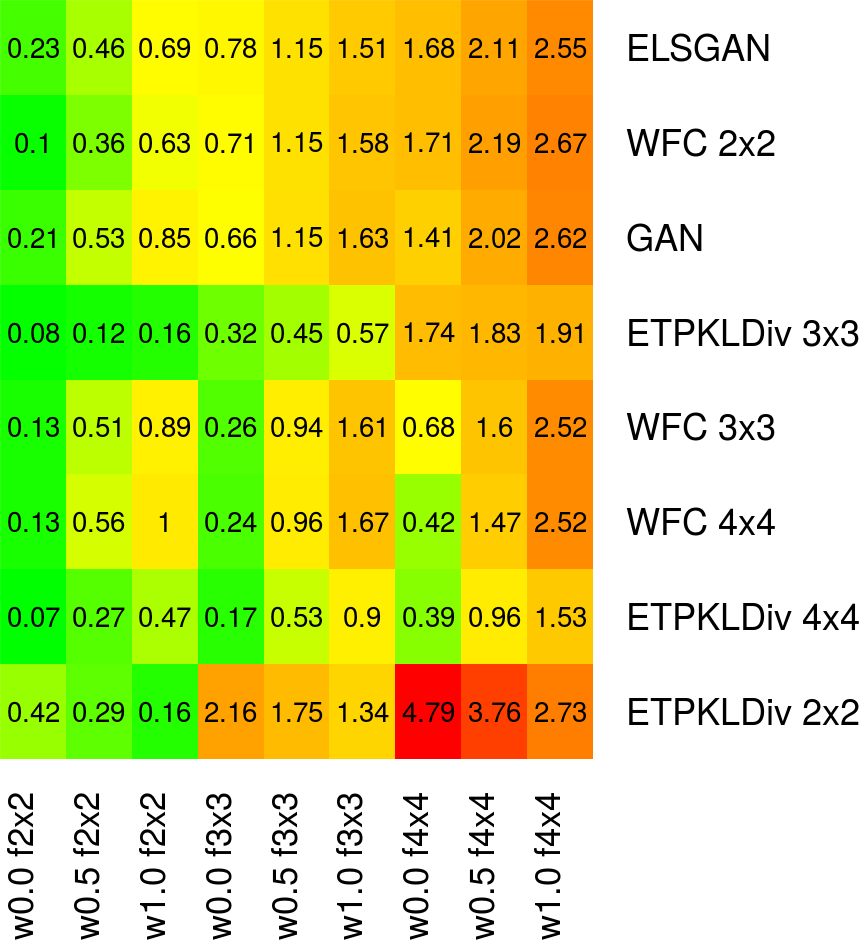}
    \caption{\label{fig:heat}Heatmap depicting mean $D_{KL}$ values for 100 levels generated by different generators}
\end{figure}

We report the mean $D_{KL}$ for each type of generator in figure \ref{fig:heat}. The parameters for the $D_{KL}$ computation are displayed on the x-axis. The respective generator is indicated on the y-axis. We display the mean value rounded to 2 digits in each cell of the resulting matrix, and visualise using colour as indicated in the colour key above the heatmap.

As expected, $D_{KL}$ values tend to be lower for small filter sizes (left three columns). However, we can also clearly see relatively good performance of ETPKLDiv with $4\times4$ and $3\times3$ filters across all $D_{KL}$ filter sizes and weights. The ETPKLDiv with $2\times2$ filter performs well for  $D_{KL}$ measures with the same filter size, but worse than all other methods on larger filter sizes. We thus conclude that the optimisation of $D_{KL}$ as suggested in section \ref{sec:kldiv:gen} was successful. However, for low filter sizes in ETPKLDiv, the successful optimisation for a smaller filter can introduce previously unseen patterns for larger filters, thus reducing the respective $D_{KL}$ value.

Interestingly, the same is not true for the WFC approaches, with all generators performing similarly across the board. This means that in the case of Mario levels, smaller filter sizes are sufficient to express constraints for placing the tiles. Both GAN approaches tend to have larger $D_{KL}$ values. This is to be expected, as the notion of KL-Divergence is slightly different. Also, the levels in ELSGAN are not optimised for $D_{KL}$, but for some other, content-specific fitness (number of jumps). We thus also see no improvement in the $D_{KL}$ values over GAN. However, in the case of ELSGAN, novel level structures are actually encouraged.

A caveat of this analysis are the large variations of the $D_{KL}$ values for most of the levels (except for ETPKLDiv). Thus, if the generator in question is fast enough, multiple levels can (and should) be produced in order to select one with the desired properties.

Something that is harder to capture is whether the methods
are predictable and reliable. Based on the rather large variances we observed, as well as from manual anecdotal observations, none
of these methods are yet reliable in that they can be 
guaranteed to produce good (as in levels which look
like the training sample and are not obviously broken)
results when given a new set of sample levels from a different game
to generate from, though they all work acceptably well for Mario levels.
They are all at the ``interesting to play with'' stage.

\section{Conclusions and future work}
\label{sec:final}

In this paper we proposed a fitness function for level generation
based on three steps:
\begin{itemize}
\item use rectangular tile patterns as features;
\item interpret the feature counts as probability distributions;
\item use Kullback-Leibler Divergence ($D_{KL}$) between generated levels and training sample levels as a measure of similarity.
\end{itemize}

The main result is that Kullback-Leibler Divergence used with tile-pattern features provides a useful and efficient way to compare game levels, providing insight into the nature of the differences. It can be applied as an objective function for evolving levels. The asymmetric nature of the measure means we can assess the novelty of the generated levels, and separately measure the extent to which features in the training set are represented in the generated levels.

The Tile Pattern KL-Div can be computed efficiently which leads to
a reasonably fast evolutionary generation process.  The levels generated
in this paper used a budget of $10,000$ fitness evaluations for the evolutionary
algorithm which took less than one second to generate levels using a $2\times2$
filter and less than two seconds when using a $4 \times 4$ filter.  Training 
time is negligible (around 1ms).

When used within an evolutionary algorithm we showed, not surprisingly, that a standard bit-flip mutation operator fails to find
fit solutions.  To counter this, we introduced a convolutional mutation operator that works by copying rectangular tile-patches from random parts of the training sample to the generated level.  This was shown to provide effective search, generating fitter levels than the samples snipped out of the training data. Our convolutional mutation operator currently always copies patches of the same size as the filter.  This is unlikely to be optimal, and we plan to introduce a bandit-based approach to sample different patch-sizes mutations.

While the method works well, it does not directly capture
capture more holistic constraints such as the way a level
may progress through different phases, and WFC also has this limitation:
any progression only happens in as much as it can be expressed
as an outcome of local tile-pattern constraints.  Our tile-pattern
implementation also has a \emph{stride} parameter to enable gaps in
the pattern which could potentially capture long-range constraints,
but we have not yet had chance to experiment with this.  The GAN
approach is able to capture long-range constraints due to the 
nature of the deep neural network, so it would be interesting to
quantify how effective this is which would involve (at least in the
case of Mario) generating and comparing wider levels.

Despite having some attractive features, the use of tile-patterns also has a significant flaw, at least in the way we have applied it here.  When sampling the tile patterns we count the number of occurrences of each exact pattern, so a pattern that differed in some small and perhaps irrelevant way is counted as being just as distinct as an entirely different pattern.  Although the method still works, a way of counting patterns that focused more on their fundamental nature and ignored irrelevant differences should work even better.  We are currently exploring a technique using connected  component analysis that offers some promise in this direction.

There are some other obvious avenues for future work:

\begin{itemize}
    \item Use an ensemble of filter sizes
    \item Apply to a wider set of games
    \item Validate measure with playtests (AI agents and human)
\end{itemize}

Finally, we believe the optimisation problem of finding levels with a low
$D_{KL}$ to be interesting in itself, and plan to propose it
as a Game Benchmark for Evolutionary Algorithms\footnote{\url{http://norvig.eecs.qmul.ac.uk/gbea/}}.
Although the method described in this paper (using our convolutional mutation operator together with
a $(1+1)$ EA) provides an acceptable solution, it is likely
that better algorithms may be developed.

\bibliographystyle{ACM-Reference-Format}
\bibliography{bibliography}

\end{document}